\documentclass[letterpaper, 10 pt, conference]{ieeeconf}  

\IEEEoverridecommandlockouts                              
\usepackage{xcolor}
\usepackage{comment}
\usepackage{graphics} 
\usepackage{epsfig} 
\usepackage{amsmath} 
\usepackage{amssymb}  
\usepackage{mathtools}
\usepackage{marginnote}
\usepackage{hyperref}

\title{\LARGE \bf
Terrestrial Locomotion of PogoX: From Hardware Design to \\Energy Shaping and Step-to-step Dynamics Based Control } 

\author{Yi Wang$^{*}$, Jiarong Kang$^{*}$, Zhiheng Chen$^{*}$, and Xiaobin Xiong   
\thanks{$^*$ The authors contribute equally to this work.}
\thanks{$^{1}$The authors are with the Wisconsin Expeditious Legged Locomotion (WELL-Lab) at the University of Wisconsin-Madison.
        {Corresponding to \tt\small xiaobin.xiong@wisc.edu}}%
}

\begin{document}
\newcommand{\bb}[1]{\mathbb{#1}}
\newcommand{\Wang}[1]{{\color{red} #1}}
\newcommand{\Kang}[1]{{\color{blue} #1}}
\newcommand{\Chen}[1]{{\color{green} #1}}

\newcommand{\block}[1]{\noindent{\textbf{#1}:}}
\newcommand{\emphhh}[1]{{\color{yellow} \textbf{#1}}}

\maketitle
\thispagestyle{empty}
\pagestyle{empty}

\begin{abstract}
We present a novel controller design on a robotic locomotor that combines an aerial vehicle with a spring-loaded leg. The main motivation is to enable the terrestrial locomotion capability on aerial vehicles so that they can carry heavy loads: heavy enough that flying is no longer possible, e.g., when the thrust-to-weight ratio (TWR) is small. The robot is designed with a pogo-stick leg and a quadrotor, and thus it is named as PogoX. We show that with a simple and lightweight spring-loaded leg, the robot is capable of hopping with TWR $<1$. The control of hopping is realized via two components: a \textit{vertical height control} via control Lyapunov function-based energy shaping, and a step-to-step (S2S) dynamics based \textit{horizontal velocity control} that is inspired by the hopping of the Spring-Loaded Inverted Pendulum (SLIP). The controller is successfully realized on the physical robot, showing dynamic terrestrial locomotion of PogoX which can hop at variable heights and different horizontal velocities with robustness to ground height variations and external pushes. 
\end{abstract}

\section{INTRODUCTION}






%
Flying robots, such as unmanned aerial vehicles (UAVs) with quadrotors as an iconic example, are increasingly popular in modern society \cite{chung2018survey}. The prevalence of UAVs can be attributed largely to the fact that they can be agile and autonomous. Such favorable characteristics have rendered UAVs with a wide range of applications, including but not limited to search and rescue \cite{schedl2021autonomous}, package delivery \cite{agha2014health}, agricultural applications \cite{elmokadem2019distributed}, and exoplanet exploration \cite{balaram2018mars}.


Despite the successes and great potential in these fields, current quadrotors have their own limitations in terms of energy consumption, short operation time, human-robot-interaction, and most importantly payload weight. Under heavy load when the thrust-to-weight ratio (TWR) is less than 1, flying is no longer feasible. Moreover, flying robots have difficulties at near ground operation and are not able to interact with the ground as they can only be stabilized in mid-air. In contrast, legged robots naturally can carry heavy loads and perform terrestrial locomotion by interacting with the ground with their feet \cite{raibert1986legged, siciliano2008springer, chignoli2021humanoid}. This motivates us to append legs to UAVs to increase their payload capacity by performing terrestrial locomotion. Thus, in this paper, we present an effective robot design by adding a light weighted spring-leg appendage to a custom designed FPV type quadrotor, namely PogoX, and, most importantly, a controller design with TWR $< 1$ to realize dynamic hopping on this robot with different horizontal velocities and hopping height.

\begin{figure}
    \centering
    \includegraphics[width=1\linewidth]{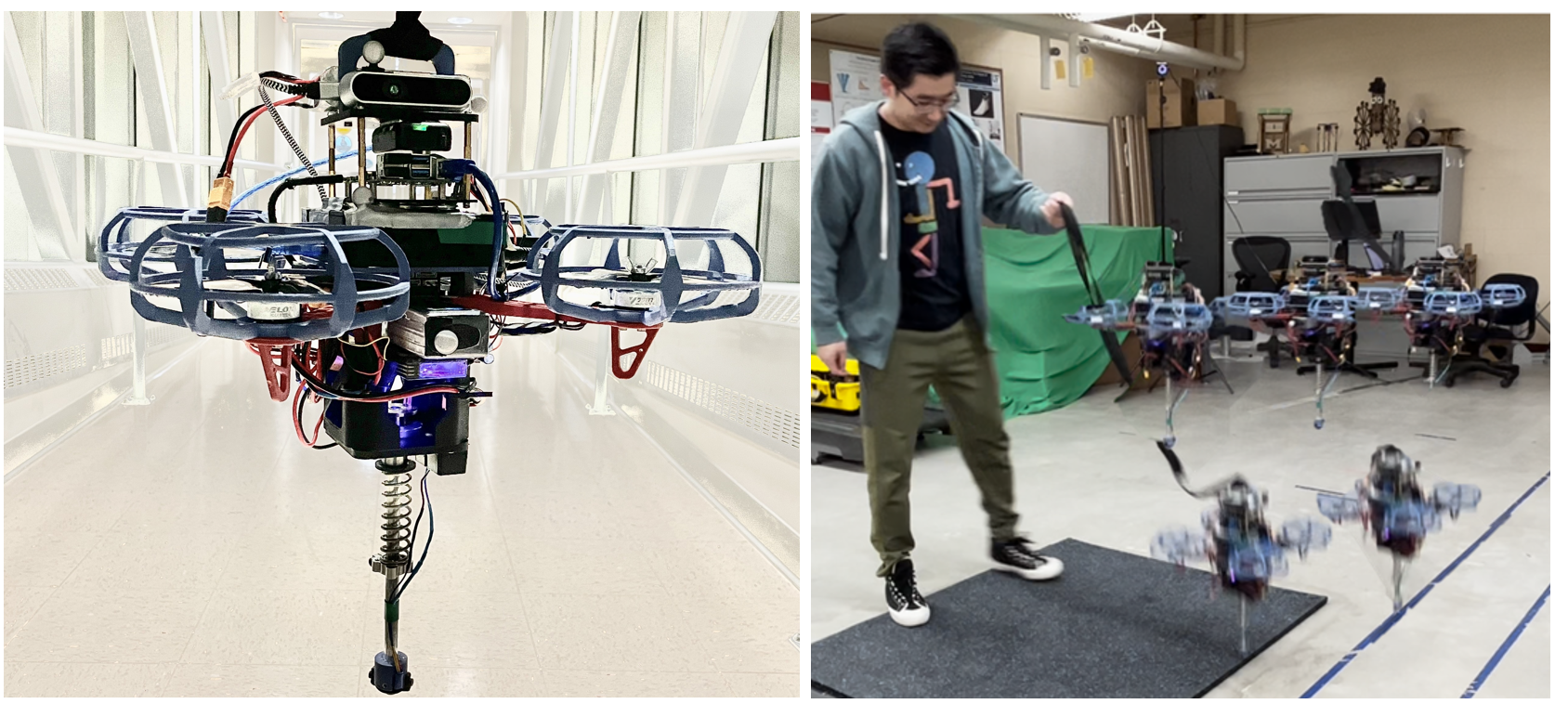}
    \caption{PogoX (left) and its hopping behavior (right).}
    \label{fig:firstfigure}
\end{figure}


Robotic systems that combine legs or wheels \cite{pratt2016dynamic,zhu2022pogodrone, kim2021bipedal,suh2020energy,  pucci2017momentum} with propellers have existed in the literature. Most work mainly focused on creating multi-modal locomotion capability rather than dynamic hopping with heavy loads, which is our focus. For instance, \cite{suh2020energy} appended wheels on drones that combine flying and wheeled locomotion. \cite{kim2021bipedal} combined a pair of bipedal legs with propellers on its torso, and \cite{pratt2016dynamic} appended a bipedal mechanism under a quadrotor; both produced flying and walking on their robots. \cite{zhu2022pogodrone} appended a spring under a miniature quadrotor that produced hopping; however the TWR of their system appears to be much bigger than one, and versatile and robust repetitive hopping behaviors have not been shown. 

With TWR $<1$, the robot becomes highly \textit{underactuated} as directly controlling the Cartesian coordinates in the continuous phases is not possible, which prevents the direct applications of existing control frameworks \cite{bouabdallah2007full, zhu2022pogodrone} on quadrotors. To address this challenge of underactuation, we draw inspiration from the legged robotic community to utilize the Spring-loaded Inverted Pendulum (SLIP) \cite{geyer2006compliant, shahbazi2016unified, hu2015torque, wieber2016modeling, xiong2018bipedal,yang2021legged, xiong2020ral} to control balancing in the horizontal directions, along with which is a Lyapunov based controller for energy stabilization in the vertical direction. 

The energy controller is driven by a control Lyapunov function-based quadratic program (CLF-QP) \cite{ames2014rapidly} to stabilize the vertical energy to a desired level, which enables periodic hybrid dynamic behaviors of hopping. Then, under this energy controller, a SLIP model is used off-line to numerically identify periodic orbits with its step-to-step (S2S) dynamics \cite{xiong20223,seipel2005running, bhounsule2014low} for feedback stabilization. The combination of the two controllers can thus realize periodic hopping on the robot with onboard computation. We realize this controller design both in simulation and hardware, showing dynamic, versatile, and robust hopping behaviors on the custom-designed robotic system of PogoX with TWR $<1$. 





\newpage

\section{Hardware System Design}
In this section, we briefly describe the hardware design of the robot, PogoX. The design architecture aims towards fully autonomous operation of this robot in the field. 
 Besides the goal of terrestrial locomotion with TWR$<$1, this design integration also can expand the operational envelope of conventional flying robots, leveraging the advantages of both flight and hopping to navigate in complex environments.

\subsection{Mechanical Design}
\begin{figure}
    \centering
    \includegraphics[width=1\linewidth]{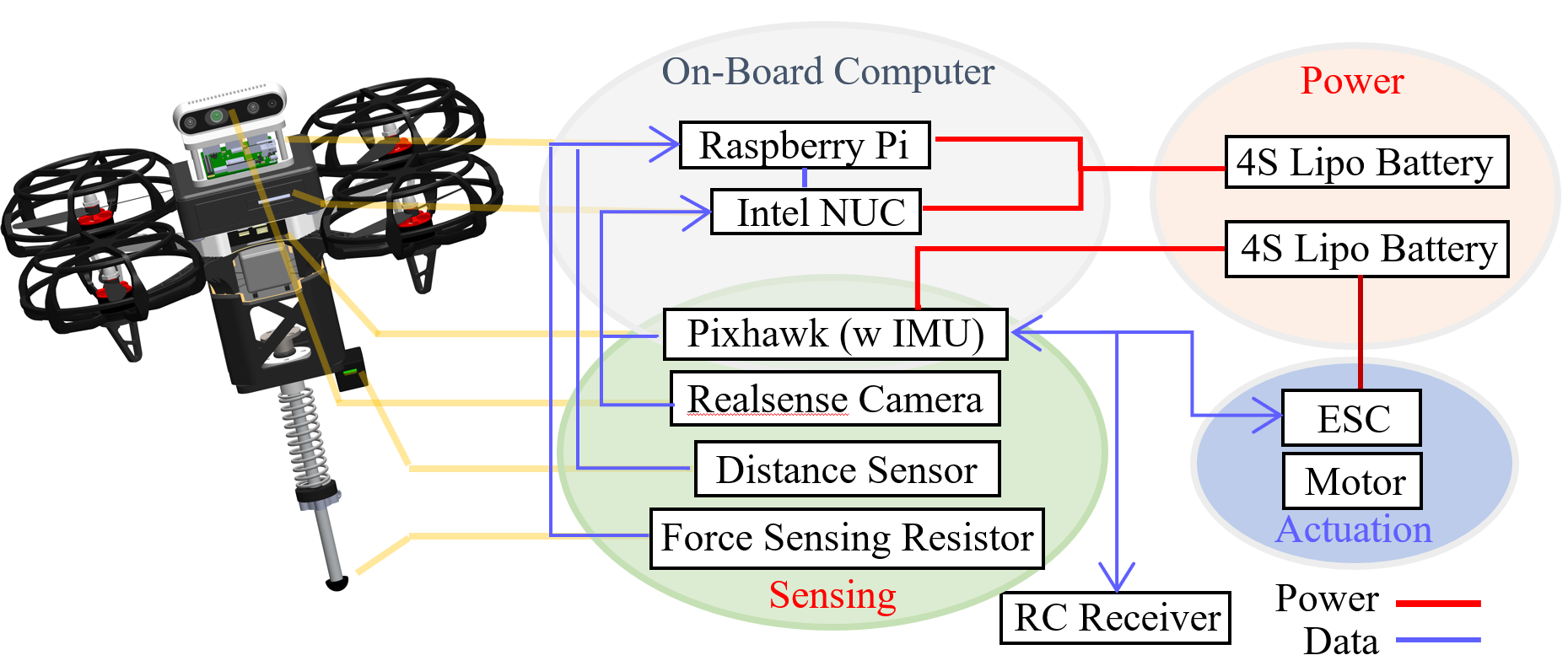}
    \caption{Illustration of the hardware system of PogoX.}
    \label{fig:hardware_system}
\end{figure}

On the mechanical design front, we select an off-the-shelf quadrotor frame that has a diagonal wheelbase at 330mm and tri-blade polycarbonate propellers that are 5.1-inch in diameter; the sizes are chosen so that various sensors, batteries, and computational resources can be mounted on the center of the body, and also that the propellers can provide sufficient thrusts. The propellers are covered by 3D printed guards to provide safety to operations with human in close proximity. The terrestrial mobility is realized via a pogo-stick leg, which can store potential energy for subsequent highly energetic maneuvers. The leg is connected through a custom-designed hollow support structure shown in Fig. \ref{fig:hardware_system}, which is made from 3D-printed PLA-CF materials. The hollow structure provides a maximum 10cm displacement for the spring. The pogo stick itself consists of an aluminum tube 
and a helical spring with a natural length of 8cm and stiffness at 4848.5N/m. Linear bearings and clamping hubs are installed in place to guide the linear motion of the leg along the body. All the components are either off-the-shelf or 3D printed. The leg appendage is of low cost and lightweight, making it straightforward to be deployed on regular quadrotors. 


 \subsection{Electronics and Control Systems}
 The electrical system of the robot is designed with the goal that the robot can perform autonomous hopping and hybrid locomotion using all onboard sensors, computational resources, and power. For the motors, we select T-Motor Velox (V2 2550KV), which has a maximum power output of 545W with a weight of 34.3g, engineered to provide the force needed for complex locomotion. A Pixhawk 6C Mini that runs the open-source PX4 software is used for motor control. To detect foot-ground contact, a FlexiForce A101 sensor is integrated with a custom-designed foot that is then mounted at the end of the leg. A Garmin LIDAR-Lite v4 LED Distance Measurement Sensor with an accuracy of ±1cm is installed on the body to detect the ground height and measure the spring displacement. We also use an Intel RealSense Depth Camera D435if to enhance perception, which in combination with an internal IMU inside the Pixhawk 6C Mini is used to provide visual-inertia odometry (VIO) for state estimation. To manage these multiple sensory inputs, a Raspberry Pi 4b is used to communicate these sensors and send the data through an Ethernet cable to an Intel NUC 13 Pro onboard computer, which performs the state estimation and optimization-based control realization. Two 4S-100C Lipo batteries with 3300mAh capacity, each weighing 12.52 ounces, are used to provide power to all the sensors, motors, and computers. With all the components installed, the robot weighs 2.5kg with a height of 0.45m. 

\section{Hybrid Dynamics of Hopping}
\label{sec:hybrid_model}

The hopping of PogoX is modeled as a dynamic system of two rigid bodies connected by a prismatic joint. The first rigid body is the floating base, consisting of the drone body, and the second rigid body is the lower part of the leg. The spring force acts directly on the prismatic joint. In total, the system has 7 degrees of freedom (DoFs), with 6 DoFs from the floating base, and another DoF from the prismatic joint. Thus, the configuration of the robot is $\mathbf{q} \in SE(3) \times \mathbb{R}$.

\begin{figure}[b]
    \centering
    \includegraphics[width=0.85\linewidth]{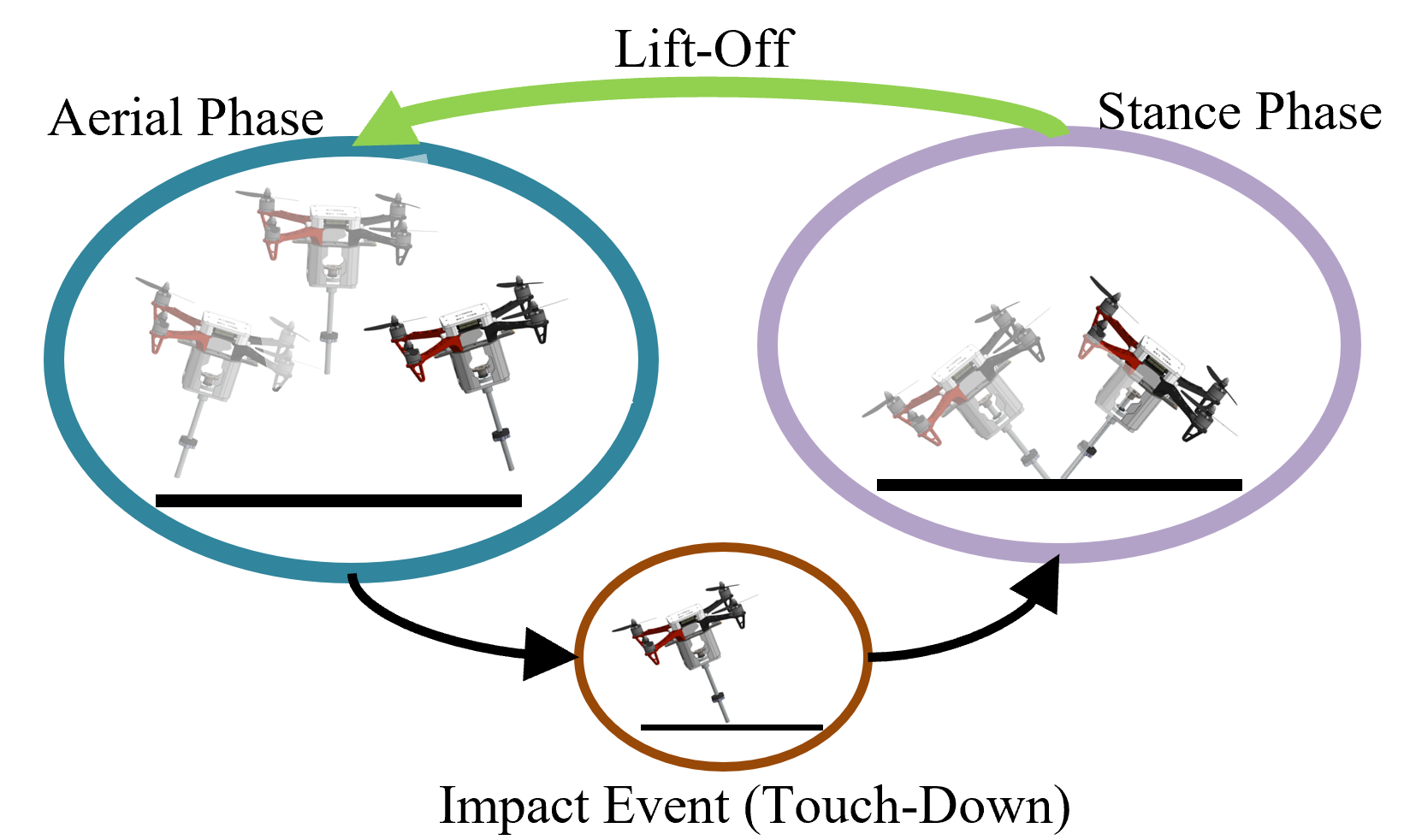}
    \caption{Illustration of the hybrid dynamics model of PogoX.}
    \label{state transition}
\end{figure}

The dynamics of periodic hopping is composed of two phases: an aerial phase and a stance phase. The transition from the aerial phase to the stance phase is a discrete impact map,  whereas the transition from the stance phase to the aerial phase is smooth. The hybrid dynamics structure is shown in Fig. \ref{state transition}.

We assume the hopping behaviors of the robot are controlled at relatively low speeds ($\le 5m/s$), thus aerodynamic drag forces are neglected. Additionally, with the presence of the leg, the propellers on the body are relatively far away from the ground, and consequently the ground effects of the propellers are considered negligible.

\subsection{Aerial Phase Dynamics}
In the aerial phase,  PogoX is actuated via its propellers. The dynamics are derived from Euler-Lagrangian Equations:
\begin{equation}
    \mathbf{M}(\mathbf{q})\ddot{\mathbf{q}} +\mathbf{H}(\dot{\mathbf{q}},\mathbf{q}) = 
    \mathbf{B}F_s+\mathbf{J}^{T}(\mathbf{q}) \mathbf{F},
    \label{manipulator equation}
\end{equation}
where $\mathbf{M}$ is the mass matrix, $\mathbf{H}$ is the Coriolis, centrifugal and gravitational forces, $\mathbf{B}$ is a constant mapping matrix, $F_s $ is the spring force, \textbf{F} represents the thrusts and moments from the propellers, and $\mathbf{J}$ is their corresponding Jacobians. 

\subsection{Stance Phase Dynamics}
In the stance phase, we assume the contact between the foot and the ground is non-slipping, which introduces holonomic constraints to the dynamics. The constraints are:
\begin{equation}
    \mathbf{J}_{f} \dot{\mathbf{q}} = \mathbf{0},
    \ \mathbf{J}_{f} \ddot{\mathbf{q}} + \dot{\mathbf{J}}_{f} \dot{\mathbf{q}} = \mathbf{0}
    \label{holonomic constraints}
\end{equation}
where $\mathbf{J}_f$ is the Jacobian of the foot position. Combining\eqref{manipulator equation} and  \eqref{holonomic constraints} yields the differential equations that govern the stance phase dynamics:

\begin{equation}
    \begin{bmatrix}
        \mathbf{M}         &   -\mathbf{J}_{f}^{T}\\
        \mathbf{J}_{f}     &   \mathbf{0}_{3 \times 3}
    \end{bmatrix}
    \begin{bmatrix}
        \ddot{\mathbf{q}}\\
        \mathbf{F}_\text{GRF}        
    \end{bmatrix} = 
    \begin{bmatrix}
        \mathbf{Bu} + \mathbf{J}^T\mathbf{F} - \mathbf{H}(\dot{\mathbf{q}},\mathbf{q})\\
        -\mathbf{\dot{J}}_{f} \mathbf{\dot{q}}
    \end{bmatrix}
    \label{stance phase dynamics}
\end{equation}
where $\mathbf{F}_\text{GRF}$ is the ground reaction force on the foot.

\subsection{Impact Mapping at Touch-Down}
We assume the impact between the foot and the ground at touch-down is purely plastic \cite{grizzle2014models}. The post-impact state satisfies the holonomic constraints, and the impulse of the impact creates a change of momentum. 
\begin{equation}
    \begin{bmatrix}
        \mathbf{M}         &   -\mathbf{J}_{f}^{T} \\               
        \mathbf{J}_{f}     &   \mathbf{0}_{3\times 3}                
    \end{bmatrix}
    \begin{bmatrix}
        \dot{\mathbf{q}}^{+}\\
        \mathbf{F}_{imp} 
    \end{bmatrix} =
    \begin{bmatrix}
        \mathbf{M} \dot{\mathbf{q}}^{-} \\
        \mathbf{0}_{3\times1}
    \end{bmatrix}
\end{equation}
where $\mathbf{F}_{imp}  $ is the impulse force from the ground, and $^+$ and $^-$ represent the states at \textit{post}- and \textit{pre}-impact, respectively.   

\subsection{Control Inputs for Hopping}
With the derivations of the dynamics in the individual domains and their transitions, we have a hybrid controlled system with the control inputs being $\textbf{F}$, i.e., the thrusts and moments of the propellers. Based on the propeller models in \cite{luukkonen2011modelling}, we assume linear relationships between the motor torque and $\omega^{2}$, and $\omega^{2}$ and propeller thrust force, where $\omega$ is the rotational speed of the propeller. Since the electric motors are controlled by ESC with sensorless Field Oriented Control (FOC), the realizable $\omega$ is in a certain range $[\omega_\text{min}, \omega_\text{max}]$. As a result, the propeller moment is bounded, i.e., $\tau \in [\tau_\text{min}, \tau_\text{max}]$ and so is the thrust force. 

With an eye towards the control of hopping, we configure the ``flight controller" of the hardware to its ``off-board" model, in which we can command desired target total thrust forces $F_t$ with desired roll, pitch, and yaw angles $[r, p, y]$; the ``flight controller" then try to realize these commands via PID controllers, the associated feedback gains of which are custom tuned. With the orientation being controlled by the flight controller, we then only need to utilize the four inputs $[F_t, r, p, y]$ to realize hopping. In order to identify the input bound of $F_t$, a linear program was formulated to solve for the minimal thrust required by the ``flight controller":
\begin{align}
   F^\text{min}_{t} &= \underset{\boldsymbol{\tau} \in \mathbb{R}^4}{\text{min}} \sum k_t \tau_i \nonumber \\
    \text{s.t.}
    &\quad \mathbf{A}_m\boldsymbol{\tau} = \mathbf{M}_b \nonumber \\
    & \quad \tau_\text{max} \geq \tau_{i} \geq \tau_\text{min}, \forall \tau_i \in \boldsymbol{\tau} \nonumber
\end{align}
where $\boldsymbol{\tau} $ represents the torques generated by all propellers, and $k_t$ is the constant ratio from the propeller moment to thrust force. $\mathbf{M_b} \in \mathbb{R}^3$ is the moment required by the flight controller represented in the body frame to realize orientation control. $\mathbf{A}_{m} \in \mathbb{R}^{3 \times 4}$ stands for the constant linear relationship between $\boldsymbol{\tau}$ and $\mathbf{M}_b$. The maximum total thrust force can be identified similarly with setting an additional constraint that $F^\text{max}_t < r mg$, where $m$ is the total robot mass, and $r < 1$ represents the thrust-to-weight ratio (TWR).

\section{Control Synthesis}
In this section, we present the controller design for realizing hopping on this hybrid system. As the orientation of the robot can be controlled directly, we mainly focus on the control of the center of mass (COM) through the hybrid behavior of hopping. The springy leg provides a desired passive dynamics of hopping. The propellers of the quadrotor can be used to inject or dissipate the energy of the whole system during hopping. Additionally, through orientation control using the propellers, the leg angles can be modulated to change the locomotion behavior. 


\begin{figure}[b]
    \centering
    \includegraphics[width=1\linewidth]{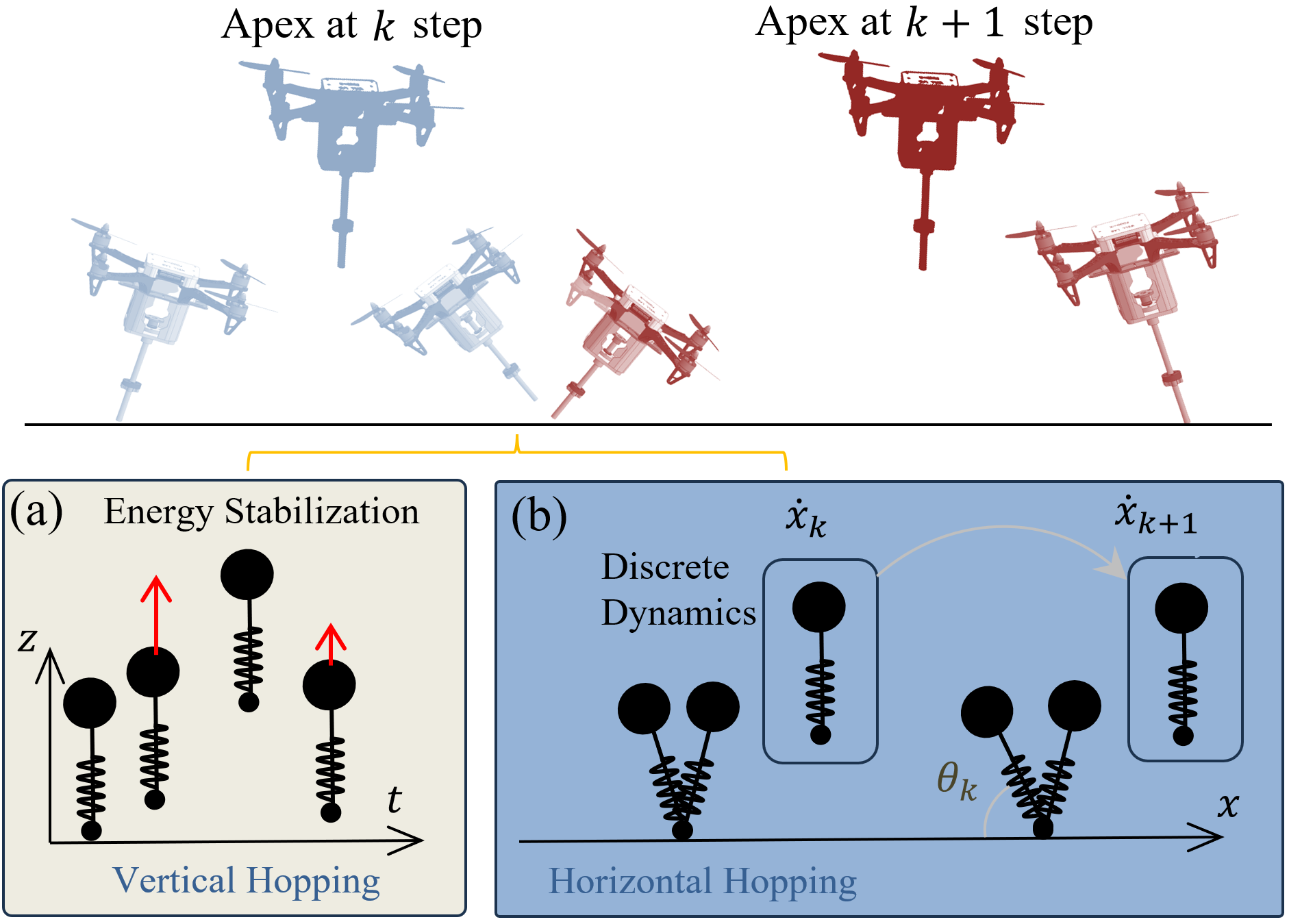}
    \caption{Illustration of the synthesis of decoupled controllers: (a) vertical energy stabilization and (b) horizontal balancing via SLIP model.}
    \label{control}
\end{figure}

The control of hopping is decoupled into a vertical energy shaping part for realizing a desired apex height, and a horizontal stabilization part for achieving a target forward velocity. In the later subsections, for the vertical control, we formulate a quadratic programs based controller to compensate for vertical energy loss due to impact and spring damping; then we describe a step-to-step (S2S) dynamics \cite{xiong2021slip, xiong2020ral, colin2023whole} based horizontal velocity controller, which is inspired by Spring loaded Inverted Pendulum (SLIP) \cite{hutter2010slip, seipel2005running} to achieve robust stepping.



\subsection{Energy Shaping for Vertical Control}


Rather than directly controlling the vertical height, we consider driving the vertical total energy to a desired value as it allows the natural flow between potential and kinetic energy during hopping. Additionally, as the ground phase is very short, we only apply the control in the aerial phase.   

Thus, the output of the control system is defined as:
\begin{equation}
    \eta  = E - E_d ,
\end{equation}
where $E_d$ is the desired constant energy level corresponding to the desired apex position. $E$ is the vertical total energy, which is the sum of the vertical kinetic energy and the potential energy coming from the gravitational force and spring force. The spring potential energy is neglected in the aerial phase. Thus $E = \frac{1}{2}m \dot{z}^2 + mg z$, where $m$ is the total mass of the system, $z$ is the vertical position,

Differentiating the output yields the output dynamics:
\begin{equation}
    \label{eq:eta_dynamics}
    \dot{\eta}  = \dot{E}  = f_{\eta} + g_{\eta} F_{t} 
\end{equation}
where $F_{t}$ is the magnitude of the total thrust force of the quadrotor. 
$f_{\eta} = 0$ in this case. 
A feedback linearizing controller can be designed for output stabilization: $ F_{t} = \frac{1}{g_{\eta}}K_p\eta, \textstyle$ where $K_p <0$ is the feedback gain. This yields a stable closed-loop dynamics:
\begin{equation}
    \dot{\eta} = g_{\eta} F_{t} = K_p \eta := A_{cl} \eta .
\end{equation}
A Lyapunov function can thus be synthesized based on the closed-loop output dynamics: $V_{\eta} = P \eta^2$, with $P>0$ satisfies the Lyapunov equation $2 A_{cl}P = -Q$, where $Q>0$ is a user-defined variable \cite{ames2013towards, khalil1996noninear}. 
Note that $\dot{V}$ is affine w.r.t. the input $F_{t}$: $
\dot{V}(\eta,F_{t}) = 2 P \eta \dot{\eta} =  2 P \eta g_{\eta} F_{t}$. Then $V$ is thus a control Lyapunov function (CLF), which motivates the use of the CLF inequality condition on the input $F_{t}$ to stabilize the output $\eta$.  The system will be exponentially stable when enforcing:
\begin{equation}
\label{eq:CLF_condition}
\dot{V}(\eta,F_{t}) \leq -\gamma V(\eta, F_{t}),    
\end{equation}
with $\gamma > 0$. This inequality is affine w.r.t. the input:
\begin{equation}
\label{eq:CLF_constraints}
    A_{\text{CLF}} F_{t} \leq b_{\text{CLF}},
\end{equation}
where $A_{\text{CLF}} = 2 P \eta g_{\eta}$, and $b_{\text{CLF}} = -\gamma P \eta^2$. 
Thus a control Lyapunov function based quadratic program (CLF-QP) can be formulated for optimizing $F_{t}$, subject to \eqref{eq:CLF_constraints} and input bounds:
\begin{align}
\label{eq:CLF_QP}
 (F_{t}, \delta) &= \underset{(F_{t},\delta) \in \mathbb{R}^2}{\text{argmin}} \  p {F_t}^2  + \delta^2\\
     \text{s.t.}& \quad A_{\text{CLF}} F_{t} \leq b_{\text{CLF}} + \delta \nonumber \\
       &  \quad F^\text{min}_{t} \leq F_{t} \leq F^\text{max}_{t} \nonumber
\end{align}
where $\delta$ is the relaxation term to ensure feasibility, and $p$ is a tuning coefficient on the cost. 


\noindent{\textbf{\emph{Remark 1}}}: In practice, $F_t$ is constantly bounded below by $F^\text{min}_t$ to avoid reversing motor rotation and maintain system operation. 
It is equivalent to adding a constant anti-gravitational force. Thus, we redefine the vertical energy:
\begin{equation}
   \textstyle E(t) = mg_ez + \frac{m\dot{z}^2}{2}, 
\end{equation}
using equivalent gravity constant $g_e = g - \frac{F^\text{min}_t}{m}$. The resulting control $F_t$ will have approximately equivalent effect to the case when the minimal bound equals 0.

\noindent{\textbf{\emph{Remark 2}}}: The inequality CLF condition in \eqref{eq:CLF_condition} enforces a convergence rate at no less than $\gamma$. Thus, the CLF-QP formulation can yield excessively large convergence rate that can lead to problematic aggressive behaviors in real systems. Inspired by task-space QP based controllers \cite{wang2020impact, xiong20223, wensing2013generation, lee2022hierarchical}, we encourage a constant convergence rate of the Lyapunov function by changing the inequality in \eqref{eq:CLF_condition} to an equality with a relaxation. The final QP becomes:
\begin{align}
\label{eq:CLF_QP_remark}
 (F_{t}, \delta) &= \underset{(F_{t},\delta) \in \mathbb{R}^2}{\text{argmin}} \ p(F_{t}-F^{pre}_{t})^2  + \delta^2\\
     \text{s.t.}& \quad A_{\text{CLF}} F_{t} = b_{\text{CLF}} + \delta \nonumber \\
       &  \quad F^\text{min}_{t} \leq F_{t} \leq F^\text{max}_{t} \nonumber
\end{align}
Minimizing the relaxation term is equivalent to minimizing the $|\dot{V} - \dot{V}^d |^2$ with the desired $\dot{V}^d  = -\gamma V(\eta, F_{t})$, which then is the cost in standard task-space control based QP formulations \cite{xiong2020ral}. Additionally, instead of penalizing the magnitude of the input $F_t$, we try to reduce the instantaneous changes of it by adding $(F_{t}-F^{pre}_{t})^2$ in the cost, where $F^{pre}$ is the applied force from the previous control sample. 

\subsection{S2S Dynamics based Horizontal Stabilization}
Now we present the horizontal state control based on a energy controlled SLIP model. The goal is to control the apex velocity in the horizontal plane. The control task is first solved in the planar case, and then applied respectively in its sagittal and lateral planes to stabilize the system in 3D.    

\block{SLIP Dynamics} We consider to use the Spring-loaded Inverted Pendulum (SLIP) model to approximate the COM dynamics of the robot, since the spring dynamics contributes heavily to the hopping behavior. Damping is added to the spring, the coefficient of which comes from the spring on the robot. Moreover, the vertical energy controller is applied on the SLIP so that the vertical energy conserves. 
Its dynamics can be compactly represented by
$
m \ddot{\mathbf{p}} = \mathbf{F}_t + \mathbf{F}_{s} + m \mathbf{g},
$
where $\mathbf{p}$ represent mass position. $\mathbf{F}_t$ and $\mathbf{F}_{s}$ are the thrust and spring forces, respectively, and they are in the same direction. 
$|\mathbf{F}_{s}| = k_s s + d_s \dot{s}$ in the ground phase with $k_s$, $d_s$, and $s$ being the stiffness, damping, and deformation of the spring, respectively, and $\mathbf{F}_{s} = 0$ with the robot is in the air. The magnitude of $\mathbf{F}_t$ is calculated from the vertical controller and its direction aligns with the leg angle. The leg is assumed massless, and thus the leg angle $\theta$ can be directly assigned in the aerial phase. For the canonical SLIP model \cite{geyer2006compliant}, $\theta$ can be set to the touchdown angle in the the aerial phase. As for our SLIP model, swing leg trajectory is part of the vertical energy-shaping controller, so a Bézier curve is used to construct a smooth swing leg trajectory $\theta(t)$ from the angle at lift-off to the desired touch down angle.

\block{Periodic Orbit and Return Map} To realize periodic hopping on the robot, we first identify periodic orbits of the SLIP. Given a desired vertical height and a target horizontal velocity, a periodic orbit with a desired touch down angle $\theta^{td}$ can be numerically identified with a given Bézier curve coefficients and the gains in the energy controller. Then we can numerically derive its return map. 

Note that for the canonical SLIP with running behavior, the dimension of the state space of the Poincaré map is reduced from four to two: the CoM height and CoM forward velocity \cite{seipel2005running, wensing2013high}. With the aforementioned vertical energy controller, our SLIP maintains a constant apex height. As a result, the Poincaré map of our SLIP has only one state variable: the CoM forward velocity $\dot{x}$. Taking the touch down angle as the input, the Poincaré map is:
\begin{equation}
\label{eq:return_map}
    \dot{x}_{k+1} = \mathcal{P}( \dot{x}_k,u_k),
\end{equation}
where $\dot{x}_{k},\dot{x}_{k+1}$ are the horizontal velocities at the apex of consecutive steps, and $u_k = \theta^{td}_k$ denotes the control input.

\block{S2S Dynamics for Stabilization}
The return map is nonlinear in nature. To utilize it for control, we derive its first-order numerical approximation:
\begin{equation}
    \dot{x}_{k+1} - \dot{x}^*= A (\dot{x}_k - \dot{x}^*) + B (u_k - u^*)+ \delta,
    \label{eq:S2S}
\end{equation}
where $A = \frac{\partial{\mathcal{P}}}{\partial{\dot{x}}}$, $B = \frac{\partial{\mathcal{P}}}{\partial{u}}$ are the Jacobian of Poincaré map evaluated at $(\dot{x}^*,u^*)$, and $\delta$ is the linear approximation error here, which will also incorporate the modeling from the SLIP to the robot. We denote \eqref{eq:S2S} as the Step-to-step (S2S) dynamics \cite{xiong20223}. The control object of driving $\dot{x}_{k+1}$ to $\dot{x}^*$ can be achieved by selecting:
\begin{equation}
\label{eq:leg_angle_controller}
    u_k = u^* + K(\dot{x}_k - \dot{x}^*),
\end{equation}
which yields the closed-loop S2S error dynamics \cite{xiong20223}:
\begin{equation}
    e_{k+1} = (A+BK)e_k + \delta
\end{equation}
with $e_k = \dot{x}_k - \dot{x}^*$. $K$ is chosen based on $(A+BK) = 0$ so that the error is stabilized in one step. In practice with $\delta$ being small and bounded, the errors converge to an error invariant set $\mathcal{E}$ \cite{xiong20223}, yielding bounded velocity tracking error.  


\section{Results}
Now we present the results of our controller implementation in both simulation and experiment. We also utilize our simulation to verify our design choices and compare energy consumption between flying and hopping for robots with different TWRs. Then, we focus on realizing different hopping behaviors on the custom-designed hardware. 

\subsection{Simulation}
The simulation is realized using MATLAB based on the hybrid dynamical model in Section \ref{sec:hybrid_model}. The dynamics are integrated via ODE45 with event triggering for domain transitions. The physical parameters of the SLIP, i.e., the mass and the stiffness, are selected to be similar to these of the robot. The periodic orbits and their S2S dynamics of the SLIP model are offline identified as a \textit{Priori} based on the selected hopping height and forward velocity. The leg angle controller in \eqref{eq:leg_angle_controller} is realized in the aerial phase to decide the target touch-down angle, which is mapped to the desired roll and pitch angles of the body. The QP-based controller in \eqref{eq:CLF_QP_remark} is realized using qpOASES \cite{Ferreau2014} at 200Hz to calculate the desired thrust force $F_t$. 
Finally, the desired control inputs of hopping $[F_t, r, p, y]$ are realized by PID controllers by sending desired rotational velocities to the propellers.
\begin{figure}[b]
    \centering
    \includegraphics[width=1\linewidth]{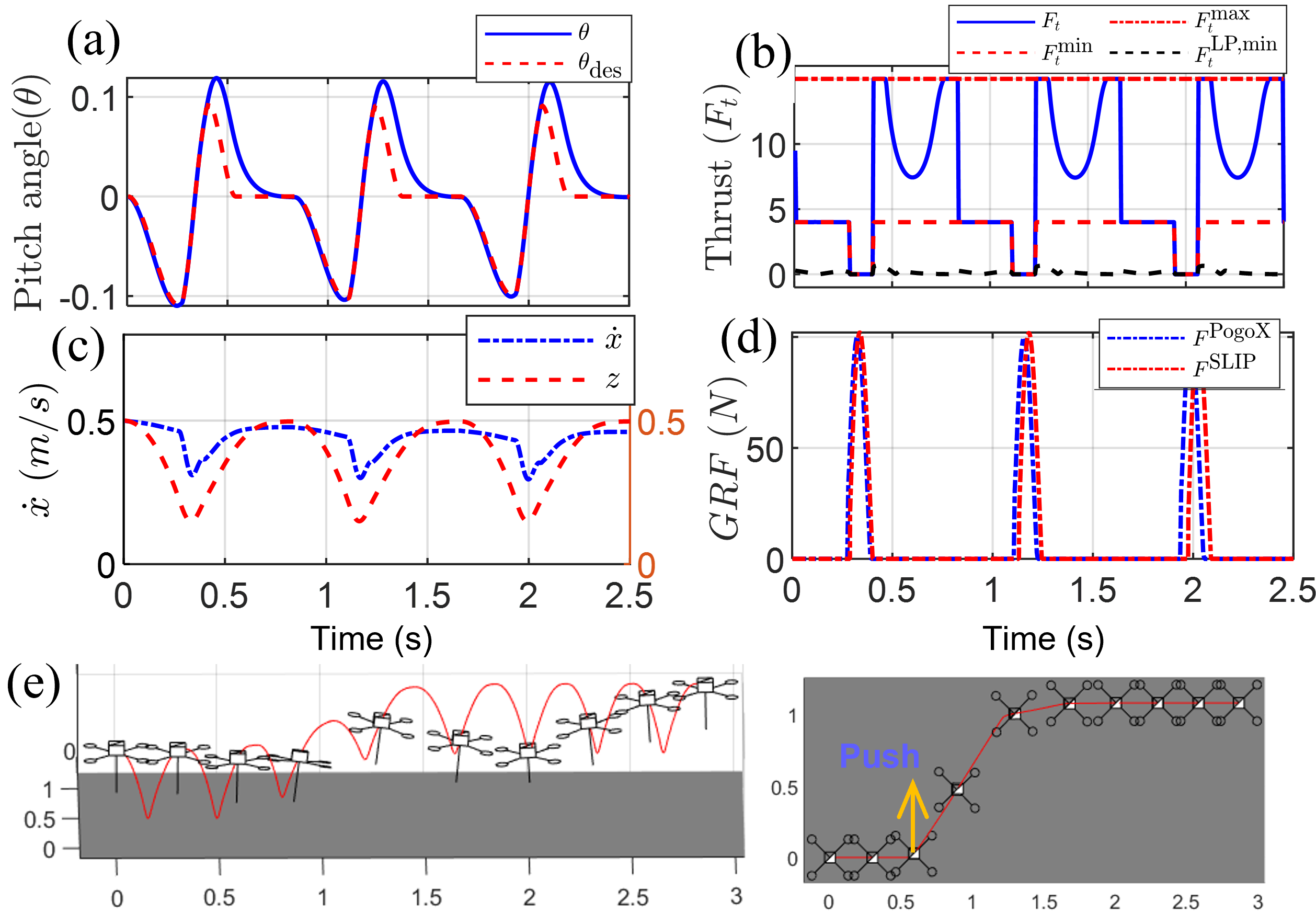}
    \caption{Simulation results: trajectories of the leg pitch angles (a), the applied thrust forces and the bounds (b), the horizontal velocity and vertical height (c), and GRFs (d) of a periodic hopping of PogoX with a desired forward velocity of 0.5m/s; (e) shows the trajectory under an external push force.}
    \label{fig:simulatioin}
\end{figure}

\block{Periodic Hopping}
We show the realized periodic hopping with a desired velocity of 0.5m/s and a desired COM apex height of 0.5m. Both the horizontal velocity and the vertical energy are stabilized as expected. We also verify that both parameters of the desired hopping behavior can be changed as we search for a different periodic orbit of the SLIP model and their associated S2S dynamics. Additionally, the implemented controller is shown to have robustness to external push disturbances. Fig. \ref{fig:simulatioin} shows the trajectories of a simulated periodic hopping of the robot, and that under an external push force at $10$N for 0.1s. 

\begin{figure}
    \centering
    \includegraphics[width=1\linewidth]{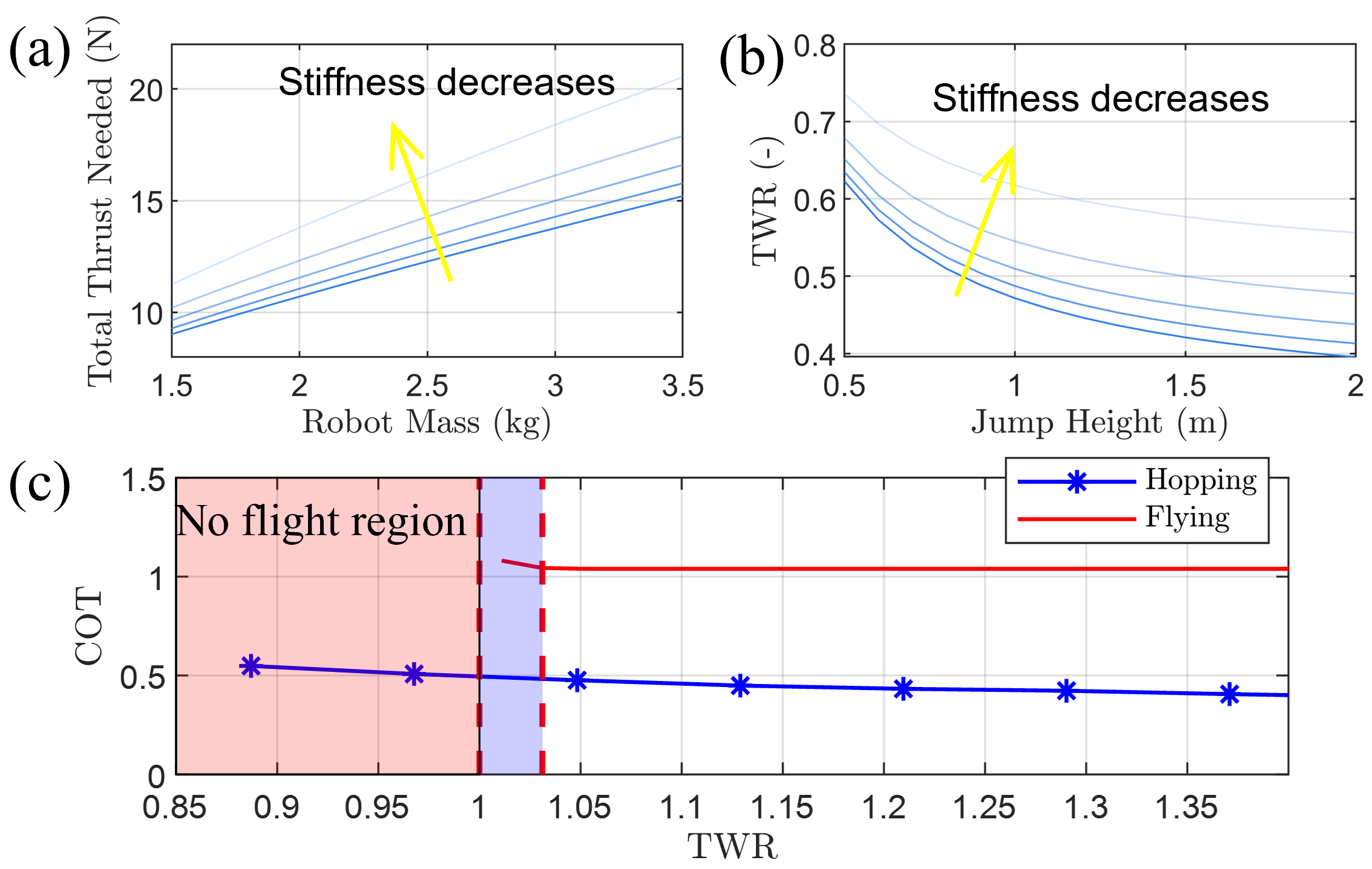}
    \caption{(a, b) Illustration of the effects of spring stiffness on periodic hopping behaviors: (a) the maximum thrust needed for different robot weights to achieve the jump height at 1.3m; (b) the TWR required to realize certain hopping height. The stiffness is increased from 2e+3N/m to 1e+4N/m with 2e+3N/m increments, and the damping coefficient is assumed to be 15Ns/m. (c) Comparison of the COT of hopping v.s. flying with a 1m/s horizontal velocity under different TWR.}
    \label{fig:design_analysis}
\end{figure}

\block{Design Analysis}
In order to obtain the design principle of system parameters (spring stiffness and maximum thrusts) with a given robot weight and target jumping height, we utilize the SLIP model to analyze the vertical hopping of PogoX. To simplify the dynamics, we assume it only hops in the vertical direction and the control is bang-bang (maximum thrust in the ascending phase and minimum thrust in the descending phase); then the dynamics become linear in all phases. Closed-form solutions are thus solved to obtain the relations between the jumping height, maximum thrust, and spring stiffness. Fig. \ref{fig:design_analysis} (a,b) shows the selection of the spring stiffness can affect the maximum thrust needed to realize a certain height for different weights, and the relation between apex height v.s. TWR.

\block{Cost of Transport} We compare the cost of transport (COT) of the same robot under different payloads between different locomotion modes: periodic hopping v.s. flying. The COT is defined as:
$
  \frac{1}{L mg}\sum\limits_{i=1}^{4}\int_{t_0}^{T_f}  \left | F_i \right | dt, 
$ where $L$ is the total horizontal distance traveled by the robot for a time period $T_f$. The energy consumption is calculated based on the integration of the applied thrust forces. Without losing generality, we control the robot to realize a 1m/s horizontal velocity in both flying and hopping; for hopping, the height is chosen to be 1m, and the minimum thrust force is set to 0 for simplification. For a fixed robot weight, we gradually decrease the maximum motor thrust forces, which equivalently decreases the TWR. Fig. \ref{fig:design_analysis} (c) shows the comparison. For flying, the COT is a finite number only when TWR $>1$. As the TWR decreases close to 1 (blue region), the controller struggles to maintain flying. At TWR $<=1$, flying is no longer feasible, producing zero velocity and thus an infinity on the COT; in reality, the TWR for realizing flying is typically chosen to be 2 or larger. For jumping, with TWR $>0.8$, the robot is able to hop to the desired height, thus producing a finite valued COT. It is thus clear that hopping is necessary with TWR $<=1$ in order to locomote, and if the horizontal velocity can be realized on hopping, hopping is more efficient than flying at TWR $>1$ as well. Note that aerial dynamics effects are neglected as we assume low-speed locomotion.


\subsection{Hardware}
The same controller is realized on the hardware. As the robot weighs 2.5kg, the maximum thrust is manually set to 22N so that TWR $<1$. The control is implemented in ROS running at 200Hz. Visual-inertia-odometry (VIO) is realized using VINS-Fusion \cite{qin2019general} to estimate the velocity and orientation of the robot during hopping. For indoor testing where the VIO alone lacks robustness due to the scenes in the environment, we use a motion-capture system to get measurements of the global position and velocity for control implementation. Since the thrust force comes from the propeller rotation, it naturally cannot deviate rapidly from the applied value at previous time stamp; we set $p=0$ in \eqref{eq:CLF_QP_remark} and the resultant QP has a closed-form solution. In the experiment, the human operator holds a tether to only provide safety for the robot when it is E-stopped. Results can be seen in the video \href{https://youtu.be/XmN6uNW69H4}{\texttt{https://youtu.be/XmN6uNW69H4}} \cite{supplementary}.

\block{Periodic Hopping} The controller is primarily implemented on the hardware to produce periodic hopping with different horizontal velocities. The desired apex height of the main body is set to 1.1m, which corresponds to a desired equivalent energy level at 10.2J. The robot is able to stabilize in place and also to desired target velocities. 

\begin{figure}[t]
    \centering
    \includegraphics[width=1\linewidth]{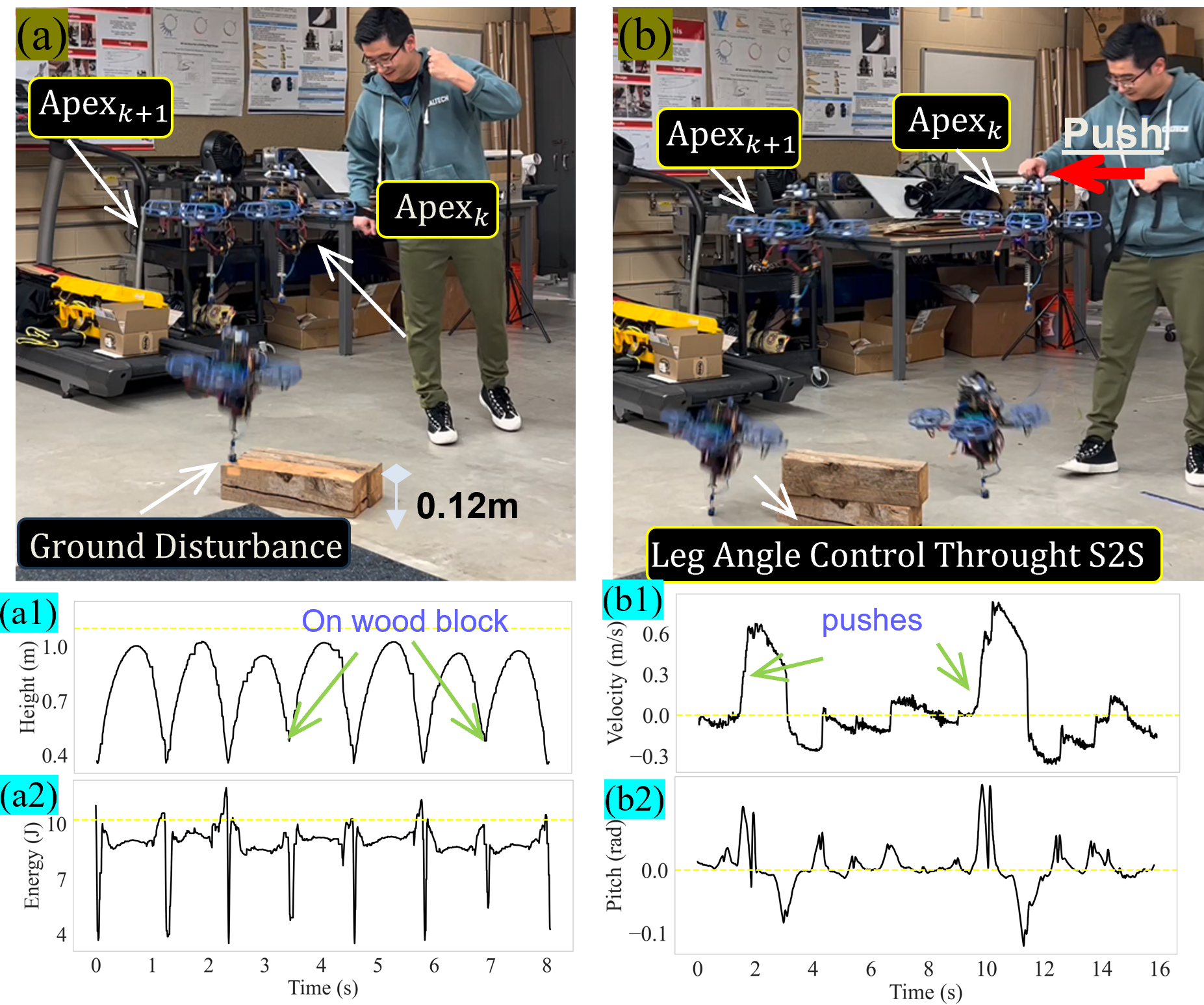}
    \caption{(a) The robot hops from the level ground onto a wood block that is 0.12m tall with the apex height remaining roughly the same: (a1) the vertical height of the robot; (a2) the energy level. (b) The robot is pushed in the air: (b1) the horizontal velocity stabilization; (b2) the pitch angle of the robot. The desired values are denoted in yellow.}
    \label{fig:robustness_z}
\end{figure}

\block{Robustness to Pushes and on Uneven Terrain} We also manually provide external pushes from the human and place wood blocks on the floor in the experiments, which represent unmodelled disturbances. It is shown that the controller can stabilize the system quickly. With the elevation from wood blocks, the hopping height remains the same due to the vertical energy controller. With the pushes, the horizontal velocities are stabilized with one step, which validates our S2S dynamics based deadbeat controller. Fig. \ref{fig:robustness_z} shows the time-lapsed images of the experiments and their associated trajectories. Both the vertical energy and horizontal velocities are stabilized with adequate robustness.

\section{Conclusions and Future Work}
To conclude, we present a controller design to enable terrestrial hopping on the robot PogoX, which is a quadrotor with a spring-leg appendage at the bottom. The key advantage of the novel system with the controller design is to enable locomotion capability on flying robots with a thrust-to-weight ratio (TWR) of less than 1. The controller is an effective combination of control Lyapunov function (CLF) driven synthesis and a Spring-loaded Inverted Pendulum (SLIP) inspired step-to-step (S2S) dynamics control. Dynamic hopping behaviors with robustness are realized in both simulation and experiment. 

In the future, we are interested in enabling PogoX to locomote outdoors in the natural environment, which will first require methodically unifying all the available advancing sensing technologies for accurate state estimation. We are also interested in exploring the optimal operations of PogoX with its flying and hopping modalities to navigate in complex and challenging environments with the goal of increasing its energy efficiency and payload capacity. 

\addtolength{\textheight}{-3.7cm}   

 \newpage
\bibliographystyle{IEEEtran}
\bibliography{main.bib}

\end{document}